\documentclass{article}

\usepackage{PRIMEarxiv}

\usepackage[utf8]{inputenc} % allow utf-8 input
\usepackage[T1]{fontenc}    % use 8-bit T1 fonts
\usepackage{hyperref}       % hyperlinks
\usepackage{url}            % simple URL typesetting
\usepackage{booktabs}       % professional-quality tables
\usepackage{amsfonts}       % blackboard math symbols
\usepackage{nicefrac}       % compact symbols for 1/2, etc.
\usepackage{microtype}      % microtypography
\usepackage{lipsum}
\usepackage{fancyhdr}       % header
\usepackage{graphicx}       % graphics
\graphicspath{{media/}}     % organize your images and other figures under media/ folder
\usepackage{subcaption}
\usepackage{graphicx} % Required for inserting images
\usepackage{minted}   % 导言区加载
\usepackage{amsmath}
\usepackage{inconsolata}
\usepackage[authoryear]{natbib}
\usepackage{listings}
\usepackage{xcolor} % For colors

\title{MooseAgent: A LLM Based Multi-agent Framework for Automating Moose Simulation
}

\author{Tao Zhang, Zhenhai Liu, Yong Xin, Yongjun Jiao \\
State Key Laboratory of Advanced Nuclear Energy Technology\\
Nuclear Power Institute of China\\
Chengdu, China\\
\texttt{taozhan22@gmail.com} \\
}

\begin{document}
\maketitle

\begin{abstract}
The Finite Element Method (FEM) traditionally involves time-consuming processes that demand specialized knowledge. This paper introduces MooseAgent, an automated system for the Moose multiphysics simulation framework that addresses these challenges. 
Leveraging Large Language Models (LLMs) and multi-agent technology, MooseAgent interprets simulation requirements from natural language to automatically generate Moose input files. To enhance accuracy and mitigate model hallucinations, the system uses a dedicated vector database constructed from annotated Moose documentation and examples.
Evaluations across typical physics cases, including heat transfer and mechanics, show MooseAgent can largely automate the simulation process, achieving an average success rate of 93\%. Furthermore, the system is highly cost-effective, with an average cost of less than 1 yuan per case. These results demonstrate the significant potential of LLMs to transform and automate complex scientific analysis. The code for the MooseAgent framework proposed in this paper has been open-sourced and is available at \href{https://github.com/taozhan18/MooseAgent}{https://github.com/taozhan18/MooseAgent}.
\end{abstract}

% keywords can be removed
\keywords{Large Language Models, Multi-agent System, LangGraph, Moose, Automated Simulation.}

\section{Introduction}
The Finite Element Method (FEM), as a powerful numerical computation tool, plays a core role in solving multiphysics problems such as structural mechanics, heat transfer, electromagnetic field analysis, and fluid-structure interaction \citep{moose, ansys}. Among these, Moose \citep{moose}, with its highly modular and flexible multiphysics coupling capabilities, has found widespread application in the nuclear energy. However, the traditional FEM simulation workflow is often complex, requiring significant time and effort from preprocessing steps like geometry creation and mesh generation, to solver parameter configuration, and post-processing stages such as results visualization and analysis. This process also demands a high level of expertise from the operator. Improper configuration at any stage can lead to simulation failure or distorted results, which undoubtedly creates a significant technical barrier, to some extent limiting the broader application and popularization of the FEM and the Moose framework.

In recent years, the field of artificial intelligence (AI) has witnessed remarkable advancements, particularly in deep learning (DL) technologies, which have achieved unprecedented breakthroughs. The integration of AI with scientific research holds the promise of accelerating simulation and design processes across various scientific and engineering domains. In the field of nuclear energy, artificial intelligence is being applied to a range of critical areas, including reactor design optimization, nuclear power plant operation and maintenance \citep{suman2021artificial,material2,huang2023review,material1}, and the research and development of nuclear fuels and materials.
Most current applications in this domain rely on machine learning or deep learning methods. However, recent advancements in large language model (LLM) technology are also demonstrating significant potential for scientific research applications within the field.

LLMs have shown immense potential in natural language understanding, knowledge reasoning, and task automation, bringing new opportunities to revolutionize scientific computing workflows \citep{lu2024ai, boiko2023autonomous, yang2024generative, tang2025matterchat}. Combining LLMs with Multi-Agent Systems (MAS) to build intelligent software agents capable of collaboratively completing complex tasks has become a research hotspot. Based on their level of encapsulation and automation, existing MAS can be broadly categorized: the first category, such as LangChain \citep{topsakal2023creating} and LangGraph \citep{chen2025implementing}, provides basic building blocks, requiring developers to code and manage agent interactions themselves, offering high flexibility but at a correspondingly higher development cost. The second category, including CrewAI \citep{venkadesh2024unlocking}, AutoGen \citep{wu2023autogen}, and MetaGPT \citep{hong2023metagpt}, offers a higher level of encapsulation where users define task objectives and agent roles, and the framework manages inter-agent communication and collaboration. The third category further lowers the barrier to entry through graphical user interfaces (e.g., Dify \citep{dify}, Bisheng \citep{bisheng}), supporting no-code workflow construction. More advanced research explores treating the multi-agent workflow itself as an optimizable object that can autonomously learn and adjust its collaboration strategies \citep{yuksekgonul2025optimizing,zhang2024aflow}.

The application of LLM-driven multi-agent systems in scientific simulation and software automation has already shown initial success. For instance, in computational fluid dynamics, researchers have experimented with using multi-agent systems to assist software like OpenFOAM, achieving an automated chain from natural language requirements to simulation execution \citep{chen2025metaopenfoam,pandey2025openfoamgpt}. The AutoFLUKA \citep{ndum2024autofluka} project utilizes LangChain to automate Monte Carlo simulations in FLUKA. These advancements fully demonstrate the great potential of LLM-driven multi-agent systems in enhancing user experience and automation levels for complex engineering software. Despite the broad application prospects of LLMs and multi-agent systems in numerous fields, related automation research in the finite element analysis domain, especially for highly complex multiphysics coupling simulation platforms like Moose, remains relatively scarce. The Moose framework offers a higher degree of freedom and more complex configuration than other finite element software, making it more difficult to master. This positions it as an ideal application scenario for introducing intelligent automation assistance, with the potential to empower more non-expert users to conveniently solve practical engineering problems using Moose.

Based on the aforementioned background and practical needs, this paper proposes an open-source automated Moose simulation multi-agent system named MooseAgent. MooseAgent employs a task decomposition strategy, combined with Retrieval-Augmented Generation (RAG) using a vector knowledge base that contains a large number of Moose input files and function documentation, and supplemented by a multi-round iterative validation and correction mechanism to generate correct Moose input files and execute calculations. In addition to significantly improving the efficiency of simulation design with Moose, MooseAgent also lays the technical foundation for developing more advanced fuel or reactor design intelligent agents in the future. This paper will systematically elaborate on the architectural design of MooseAgent, its core algorithmic workflow, the construction details of its specialized knowledge base, and will comprehensively evaluate its performance through a series of typical simulation cases. The code for the proposed MooseAgent is open-source and available at \href{https://github.com/taozhan18/MooseAgent}{https://github.com/taozhan18/MooseAgent}.

\section{Method}
\subsection{Overall archetecture}
\begin{figure*}
	\centering
	\includegraphics[width=0.5\columnwidth]{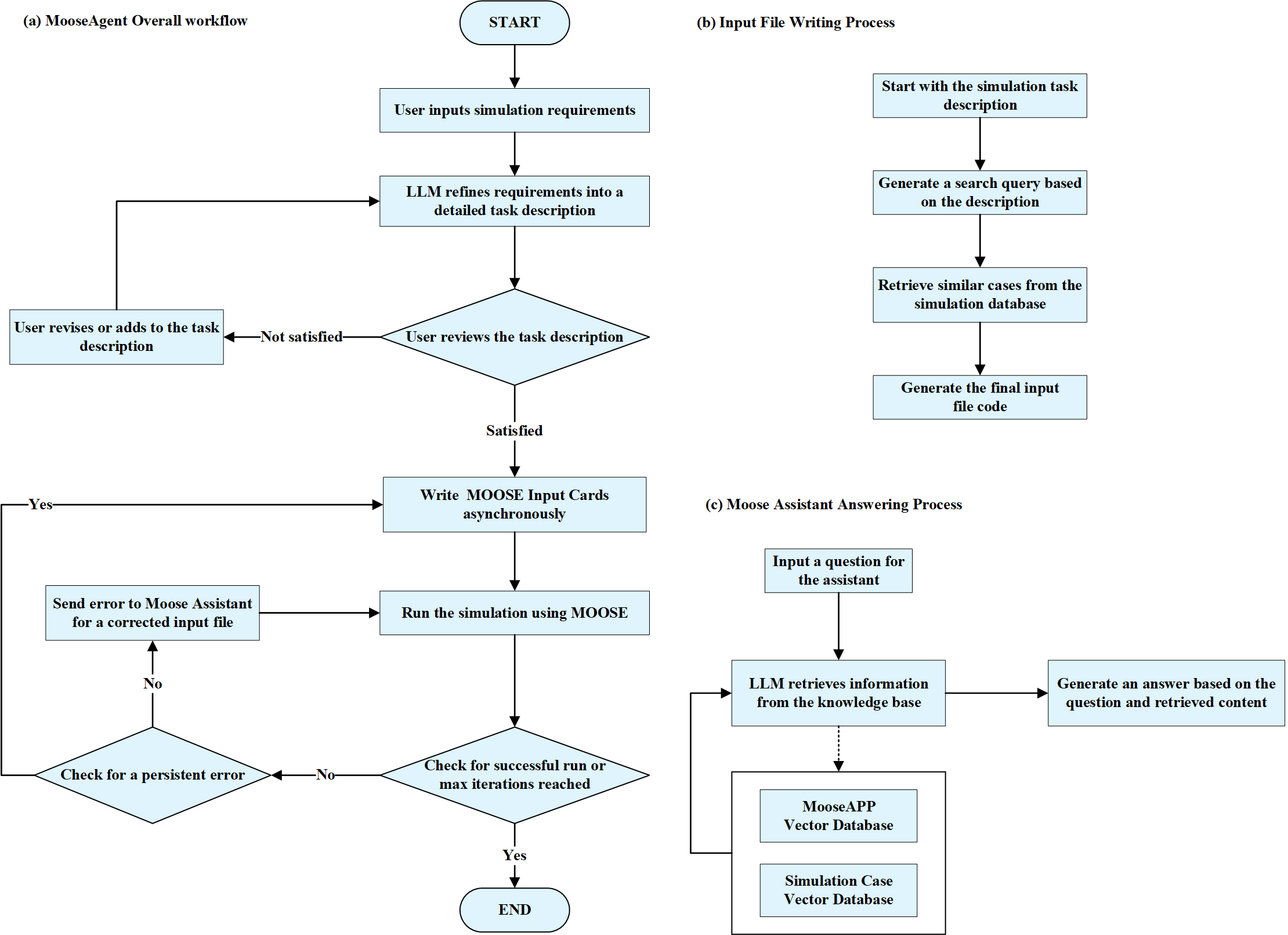}
	\caption{Overview of MooseAgent.}
	\label{FIG:architecture}
\end{figure*}

The overall framework of MooseAgent is shown in Fig.\ref{FIG:architecture}. Its core idea is to leverage the natural language understanding capabilities of LLMs and the collaborative problem-solving abilities of MAS to achieve an automated workflow from user requirements to Moose simulation results. The entire framework is built on LangGraph, and we assign specific roles to each agent and precisely define their interaction sequences, collaborative logic, and the information passed between them. This ensures efficient task execution and robust error handling.
% 整个框架基于langGraph搭建，为每个智能体分配特定角色，智能体之间地交互顺序和逻辑，定义智能体之间传递地信息，确保任务高效执行与错误妥善处理。

The framework can be divided into three primary parts, each with specific responsibilities and actions:

\textbf{Alignment}: This part generates a detailed simulation task description based on the user's requirements and aligns the simulation goals through interaction with the user.

\textbf{Write Input Card}: This part asynchronously writes the MOOSE input file based on the detailed simulation task description.

\textbf{Execute-Analyze-Correct}: This part corrects errors in the input file through multiple rounds of iteration to ensure it runs correctly.

Each core function is executed by a dedicated agent. Every agent operates based on a unique set of prompts, which consists of a 'System Prompt' and a 'User Input Prompt'. The prompt design for the alignment agent is illustrated in Appendix.\ref{sec:prompt}.
% 每个部分都有相应的智能体完成其功能，每个智能体都有与其适配的提示词，分为系统提示词和用户输入提示词。图中给出了对齐智能体的提示词设计，更多智能体的提示词可见代码仓库，

% \begin{figure}
% 	\centering
% 	\includegraphics[width=0.5\columnwidth]{FIG_P.png}
% 	\caption{Prompts of MooseAgent.}
% 	\label{FIG:prompt}
% \end{figure}

First, the user inputs the simulation requirements in natural language. The agent then begins by deeply understanding and parsing these requirements, which may be vague or unclear. It determines the number of Moose input files needed and the specific tasks each file must accomplish. During this process, the LLM may engage in a multi-turn dialogue with the user to clarify any ambiguous descriptions, thereby ensuring the accuracy of the task definition.
Subsequently, the system asynchronously writes the Moose input files based on the task descriptions. For tasks requiring multiple input files, asynchronous execution enhances the efficiency of code generation. In the writing process, the system first formulates a search query based on the simulation task description. It then retrieves similar input files from a simulation case database to serve as references for generating the initial input file code.

Once an input file has been written, a multi-round "execute-analyze-correct" iterative process begins. If an error or warning is encountered during the Moose execution, the error information is returned. The agent first analyzes the error history to determine if it is encountering a recurring error. If so, it regenerates the input file and suggests a different implementation approach to avoid the error. Otherwise, it calls upon a "Moose Assistant" for a specific solution. The "Moose Assistant" is an LLM configured with a Moose knowledge base, enabling it to selectively retrieve relevant content to answer technical questions about the Moose simulation. This iterative process continues until the simulation runs successfully or a preset maximum number of iterations is reached.

\subsection{database}
To reduce the "hallucination" phenomenon in LLMs—where the model may generate inaccurate or irrelevant content—and enhance the accuracy of generated content, this paper constructs a specialized Moose vector database composed of two core parts.

\begin{figure}
	\centering
	\includegraphics[width=0.5\columnwidth]{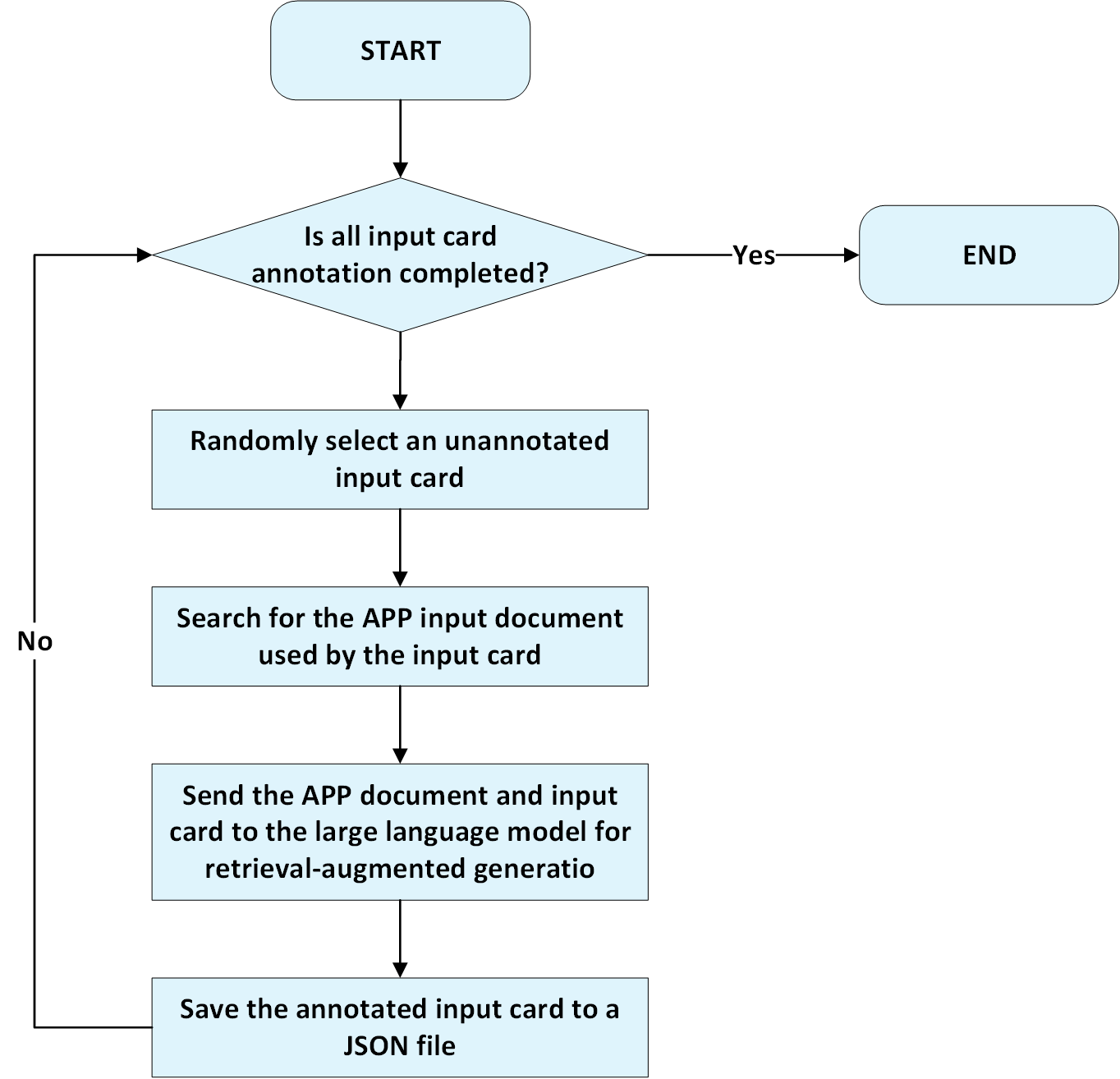}
	\caption{Automatic annotation workflow.}
	\label{FIG:COMMENT}
\end{figure}

The first part consists of annotated Moose input files. Over 8,000 input file instances were collected from Moose's official code repository, primarily sourced from various test cases and tutorials. However, the original input files largely lack detailed comments. Using them directly for retrieval could make it difficult for the model to understand their true intent or lead to misuse.
To address this, an automated input file annotation workflow was designed and implemented, as shown in Fig.\ref{FIG:COMMENT}. This workflow involves the following steps:
\begin{itemize}
    \item First, it randomly selects samples from the unannotated input files.
    \item Next, by analyzing the types of Moose functions used, it retrieves descriptions of their functionality, usage methods, and relevant parameter details from the Moose documentation.
    \item Finally, using RAG technology, it leverages the retrieved contextual information to drive an LLM to generate detailed comments for the input file.
This annotation process is executed iteratively, continuously expanding the library of annotated input files. All annotated input files are stored in JSON format, with each entry containing the input file's name, a summary of its simulation task, and the input file code with detailed comments.
\end{itemize}
% 数据格式例子
The second part is the detailed documentation for all functions included in Moose. While Moose has a powerful documentation system that can export the basic usage of all functions via commands, it often lacks descriptions of specific functionalities and application scenarios.
To address this, a Python script was developed to scan the Moose documentation library, locating and extracting detailed textual descriptions and usage explanations for each function and its core features. The extracted information is organized and stored in JSON format, with each entry containing the function's name, a comprehensive description of its functionality, and detailed explanations of its input parameters, including data types and default values.
An example of the database can be seen in Appendix.\ref{sec:data}.

% \begin{figure}
%   \centering
%   \subcaptionbox{Annotated Moose input file\label{FIG:inputcard}}
%     {\includegraphics[width=1\linewidth]{CARD.png}}
%   \subcaptionbox{Moose function\label{FIG:function}}
%     {\includegraphics[width=1\linewidth]{APP.png}}
%   \caption{Database of MooseAgent.}
%   \label{FIG:database}
% \end{figure}

Together, these two sources constitute a comprehensive knowledge base that serves as the foundation for MooseAgent. To make this information accessible, we employ the state-of-the-art BGE-M3 embedding model \citep{chen2024bge} to encode the entire corpus of textual data into semantically rich, high-dimensional vectors. These vectors are then stored and indexed in a FAISS vector database \citep{douze2024faiss}, a system chosen for its exceptional efficiency and ease of use. FAISS is particularly well-suited for fast similarity searches and indexing of dense vectors. For any given query, the system calculates relevance using a similarity score and selects the top 3 chunks with the highest scores. The retrieved information is then combined with the user's original message using a simple stacking method, which creates an enriched context. This method involves layering the retrieved chunks directly with the user's query to provide a comprehensive context for the LLM.
Through the above technologies, we effectively provide the LLM with accurate domain knowledge, thereby significantly enhancing the accuracy, reliability, and overall success rate of the entire automated workflow.
\section{Experiments}
\subsection{setting}
To optimize performance, we have adopted a specialized approach. The more powerful Deepseek-R1, with its superior reasoning capabilities, is dedicated to the core task of generating the input file. Meanwhile, the Deepseek-V3 model handles the remaining modules, ensuring high-quality output while accelerating the overall process. The temperature, a key LLM parameter that controls the randomness of the generated text, was set to 0.01 to ensure highly focused and deterministic outputs. We set the maximum number of iterations to three. Through testing, we found that a higher number of iterations was not beneficial, as it typically indicated that the problem exceeded the model's problem-solving capabilities.

To evaluate the framework's effectiveness, we conducted experiments using nine distinct test cases, each repeated five times. Performance was measured by success rate and token consumption. The cases included:
\begin{itemize}
    \item \textbf{Steady-State Heat Conduction:} Simulates the steady-state temperature distribution in a one-dimensional metal rod with constant temperatures at both ends.
    \item \textbf{Transient Heat Conduction:} Simulates the transient evolution of the temperature field over time in a two-dimensional square region under specific initial and boundary conditions.
    \item \textbf{Linear Elasticity:} Simulates the linear elastic stress and strain fields within a two-dimensional rectangular plate under fixed and loaded boundary conditions.
    \item \textbf{Plasticity:} Simulates the final plastic strain distribution in a two-dimensional square plate after the material enters an elastoplastic state under fixed and displacement loading conditions.
    \item \textbf{Porous Media Flow:} Calculates the steady-state seepage field and pressure distribution in a three-dimensional cubic soil block under set hydraulic head boundary conditions.
    \item \textbf{Phase Change Heat Conduction:} Simulates the movement of the phase change front during the melting/solidification process of a one-dimensional metal rod driven by a temperature difference at its ends.
    \item \textbf{Phase Field:} Uses a phase-field model to simulate the solidification process of a two-dimensional pure metal driven by low temperature, observing the formation, growth, and interface evolution of the solid phase.
    % 注意：这里的 ** 已经被修正为 :
    \item \textbf{Prismatic Fuel:} Loads a predefined 16-sided prismatic fuel assembly mesh, sets temperature and mechanical boundary conditions, and solves for the fuel's thermal and mechanical performance.
    \item \textbf{Thermal-Mechanical Coupling:} Utilizes Moose's Multiapp functionality to simulate the thermal-structural coupled behavior of a two-dimensional rectangular thin plate, analyzing thermal expansion and displacement.
\end{itemize}

\subsection{results}
Table~\ref{tab:performance} summarizes the performance of MooseAgent across 9 different test cases, achieving an impressive average success rate of 93\% over all cases. MooseAgent attained a perfect success rate (100\%) in 7 out of the 9 scenarios: steady-state and transient heat conduction, linear elasticity, phase change heat conduction, porous media flow, phase field, and prismatic fuel. The thermal-mechanical coupling case also demonstrated high reliability with an 80\% success rate. This strong performance indicates that for problems where the physical models are relatively mature and the solution process is fairly standard, MooseAgent can accurately understand user requirements and generate correct input files, likely benefiting from a knowledge base rich with similar reference cases. 
The case with the lowest success rate was plasticity, at 60\%, which indicates that the performance of the agent in dealing with the complex nonlinear behavior of materials still needs to be improved.
In terms of token usage, there were significant differences between cases, mainly related to task difficulty. The average token consumption was approximately 60,777 tokens. If calculated based on the pricing for calling the DeepSeek-R1 model, this would cost less than 1 yuan per run. Therefore, MooseAgent can perform simulation tasks at a very low cost.

\begin{figure}
	\centering
	\includegraphics[width=0.8\columnwidth]{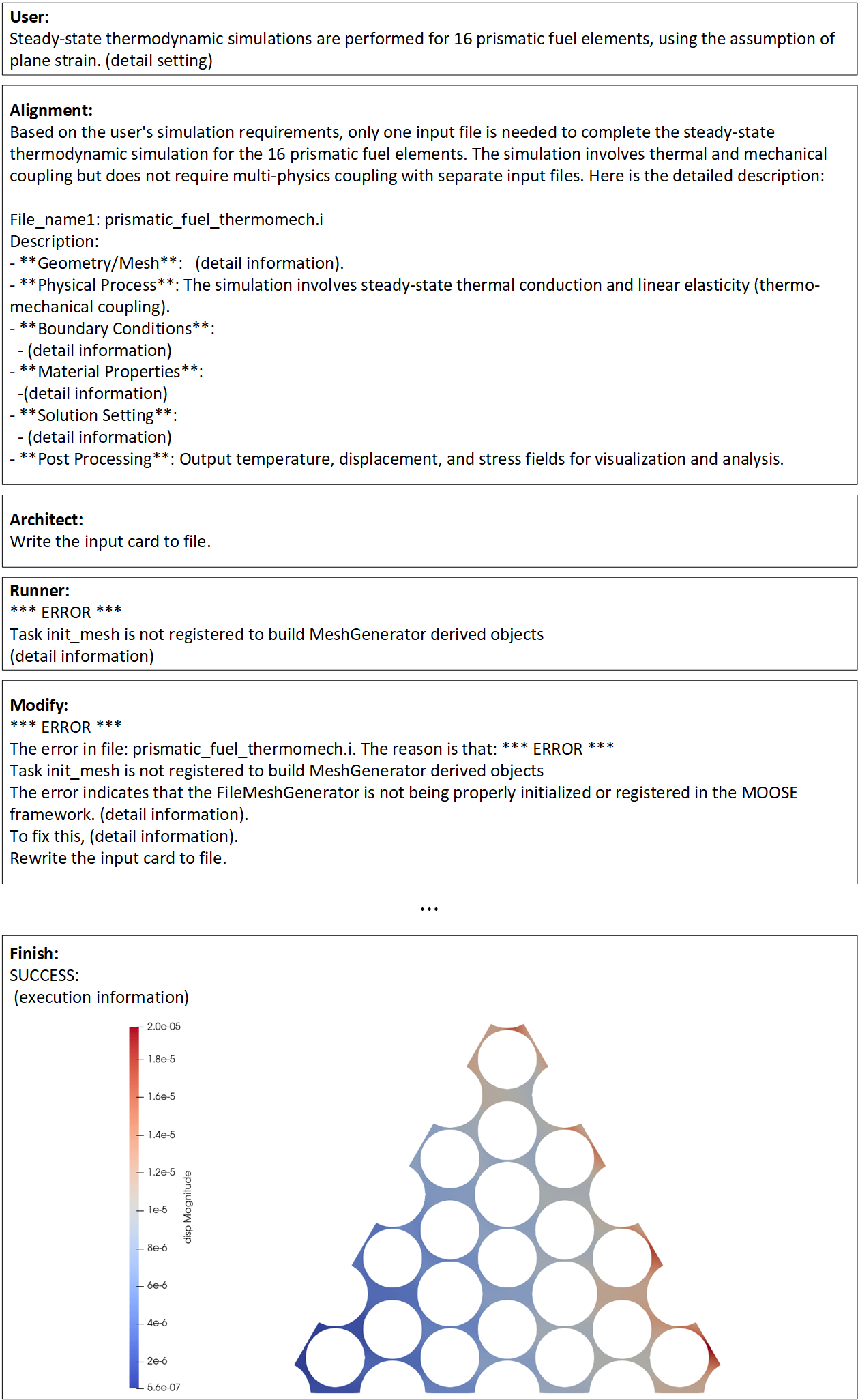}
	\caption{Prismatic fuel element.}
	\label{prismatic fuel element1}
\end{figure}

Fig.~\ref{prismatic fuel element1} illustrates a typical MooseAgent workflow using a prismatic fuel element simulation as an example. Initially, the user describes the basic requirements for the thermal-mechanical simulation in natural language, specifying details such as the names of boundaries and blocks in the mesh file and their corresponding boundary conditions. The LLM then comprehends and refines these requirements, generating a detailed description that encompasses the mesh, physical processes, boundary conditions, material properties, and solver settings. This description is displayed on the console for user confirmation. Once the requirements are confirmed, a complete input file is generated and executed. If MOOSE returns an error during execution, a multi-turn iterative correction process is initiated. In this process, the LLM analyzes the error message and source code to identify the cause, then modifies the input file for re-execution. Ultimately, the simulation runs successfully, yielding contour plots that show the distribution of the fuel's thermal-mechanical performance.

An analysis of failed runs reveals that all unsuccessful attempts were caught in an infinite loop. Specifically, the agent would get stuck repeatedly trying to fix an error by calling a particular function. Because it could never successfully execute the function call, it would eventually exit after reaching the maximum number of iterations.
The root cause of this issue is twofold. First, as an open-source software, MOOSE has a vast number of available functions, making it difficult for the LLM to select the most appropriate one for a given task. Additionally, the LLM lacks accuracy in handling some of MOOSE's fundamental functionalities. Specifically, when writing function expressions, understanding the basic format of input cards, defining constants, and establishing relationships between different modules, it occasionally falls into a loop of irreparable errors.
To address this, future improvements could focus on two key areas. One approach is to incorporate a human feedback mechanism to guide the model on how to escape its current error loops. Another strategy is to fine-tune the large model to enhance its understanding of specific MOOSE conventions. Both methods are expected to increase the overall task success rate.

\begin{table}[h!]
\centering
\caption{Performance of MooseAgent}
\label{tab:performance}
\begin{tabular}{lcc}
\hline
\textbf{Case} & \textbf{Success Rate} & \textbf{Tokens} \\
\hline
Steady-State Heat Conduction & 1 & 24673 \\
Transient Heat Conduction & 1 & 26729 \\
Linear Elasticity & 1 & 40857 \\
Plasticity & 0.6 &  82180\\
Phase Change Heat Conduction & 1 & 64592 \\
Porous Media Flow & 1 & 104766 \\
Phase Field & 1 & 86696 \\
Prismatic Fuel & 1 & 37424 \\
Thermal-Mechanical Coupling & 0.8 & 79075 \\
\hline
\end{tabular}
\end{table}

\subsection{Ablation experiments}
To evaluate the effectiveness of the RAG approach, we conducted an ablation study by running the same test cases without the knowledge base. The results, presented in Table~\ref{tab:ablation}, show a significant decline in performance. The overall average success rate dropped from 93\% to 76\%.

This decrease was most pronounced in the more complex cases. For instance, the success rate for the plasticity problem plummeted from 60\% to just 20\%. Similarly, phase change heat conduction and thermal-mechanical coupling saw their success rates fall from 100\% to 60\% and from 80\% to 60\%, respectively. In contrast, the simplest cases, such as steady-state heat conduction and the prismatic fuel element, maintained a 100\% success rate, indicating that these tasks are simple enough for the base model to handle without specific examples.

Interestingly, the average token consumption was slightly lower in the ablation study, with 54,724 tokens used compared to 60,777 in the full study. This reduction is attributed to the omission of certain input card and function documentation sections in the input provided to the large model, which significantly decreased the number of tokens per input. However, it was also observed that token consumption increased for some tasks, such as transient heat conduction, plasticity, and thermo-nuclear coupling problems. This increase is likely due to the absence of guidance from the knowledge base, which necessitates additional iterative corrections to generate accurate input cards. 
In summary, this experiment confirm that the knowledge base is a critical component, providing essential context and examples that significantly enhance the agent's ability to solve complex, non-linear, and multi-physics problems accurately.
\begin{table}[h!]
\centering
\caption{Performance of MooseAgent without database}
\label{tab:ablation}
\begin{tabular}{lcc}
\hline
\textbf{Case} & \textbf{Success Rate} & \textbf{Tokens} \\
\hline
Steady-State Heat Conduction & 1 & 11194 \\
Transient Heat Conduction & 1 & 33871 \\
Linear Elasticity & 0.8 & 22445 \\
Plasticity & 0.2 & 100590 \\
Phase Change Heat Conduction & 0.6& 35241\\
Porous Media Flow & 0.8 & 77502 \\
Phase Field & 0.8 & 97993 \\
Prismatic Fuel &  1 &  19216\\
Thermal-Mechanical Coupling & 0.6 &  94468\\
\hline
\end{tabular}
\end{table}
\section{Conclusion}
This paper proposes an automated Moose simulation multi-agent system named MooseAgent. The system can understand user requirements in natural language and combines task decomposition, RAG, and a multi-round iterative validation and correction strategy to automatically generate Moose input files and perform simulations, thereby lowering the barrier to entry for Moose and improving simulation efficiency. Experimental results show that MooseAgent achieves a high accuracy rate on problems with relatively simple physical processes and can also succeed on more complex problems through multiple attempts, with an average success rate of 93\% across all cases. Furthermore, the token consumption for all cases is kept at a low level, with the average cost of running a single simulation task being approximately less than 1 yuan, demonstrating its high cost-effectiveness.
In addition, further ablation experiments are conducted to demonstrate the effectiveness of database and RAG.
Future work will focus on optimizing knowledge retrieval and iterative strategies, and will consider introducing human feedback into the iteration process to further enhance the system's performance on complex problems.

%Bibliography
\appendix
\section{Prompt of MooseAgent}
\label{sec:prompt}
This Appendix will present the prompt for alignment. For the complete prompts for all agents, please refer directly to our \href{https://github.com/taozhan18/MooseAgent}{repository}.
The prompt is divided into a system-level prompt and a user-level prompt. The system-level prompt is a higher-priority instruction that defines the agent's role, the task to be completed, and the overall requirements. In contrast, the user-level prompt provides a detailed breakdown of the specific analysis, including the main modules, the analysis to be performed on the MOOSE simulation, and the output requirements.

\begin{figure*}
\begin{lstlisting}[]
SYSTEM_ALIGNMENT_PROMPT = """Your task is to supplement the details that are not mentioned but need to be set for calculation based on the simulation requirements provided by the user, for the user's confirmation. Ensure detailed and comprehensive descriptions, and more importantly, set deterministic and quantitative descriptions to avoid vague statements.
"""
HUMAN_ALIGNMENT_PROMPT = """
The following are the user's simulation requirements:
<simulation_requirement>
{requirement}
</simulation_requirement>
Here is the feedback from the user (if any):
<feedback>
{feedback}
</feedback>
You first need to determine how many input files need to be built to complete the simulation task (one is sufficient for most problems, but multi-physics coupling problems usually require multiple), and determine the name (including suffixes) of each input file, as well as the purpose of each input file. For each input file, you need to provide a detailed description. If it is an .i file, it usually needs to include geometric shapes, physical processes, boundary conditions, solution settings, etc., as follows:
-Geometry/Mesh: Clarify the geometric features such as shape and structure of the simulated object. Provide specific and quantitative grid partitioning methods to determine the dimensions (1D, 2D, 3D) and coordinate systems (RZ, Cartesian, etc.) of the problem.
-Physical Process: Describing the physical phenomena and principles involved in simulation, such as mechanics, thermodynamics, electromagnetics, etc.
-Boundary conditions: describe the boundary conditions on the boundary of the simulation area, such as the action of forces, temperature distribution, and the type of boundary conditions (Dirichlet, Neumann, etc.)
-Material: Describe the material properties of the simulated object that need to be used in the simulation, such as density, elastic modulus, Poisson's ratio, etc.
-Solution setting: Determine the solution method, time step, convergence criteria, etc. used in the simulation.
Finally, prompt the user to confirm whether the simulation description meets their requirements.
-Transfer and MultiApps (if any): If the input card has sub-card, it is necessary to define MultiApps and Transfers. Within MultiApps, the file names to be transferred should be specified, and within Transfers, the variable names required for the transfer should be defined. If the input card only needs to receive variables from other input cards, there is no need to define MultiApps and Transfers within the input card. Instead, the variable names should be specified here. If it is outside of these two situations, the module does not need to be defined.
-Post processing (if any): Describe the post-processing methods and results of the simulation, such as the distribution of temperature, stress, displacement, etc.
You should reply a like this (Please note that if there are multiple input cards, make sure the first file is the main card used for execution. Please note that when using Transfer and MultiApps, please check the consistency between the sub-card names and corresponding file names, and ensure that the definitions of the transferred variables are consistent across different input cards. In addition, loop calls within MultiApp should be avoided.):
<write the number of files> needed to complete the simulation task, the main card is: <the name of main card>.
File_name1: Write the file name.
Description: Write the detailed description of the file.
File_name2: Write the file name.
Description: Write the detailed description of the file.
...
"""
\end{lstlisting}
% \caption{Prompts of MooseAgent.}
% \label{FIG:prompt}
\end{figure*}

\section{Moose Database}
This Appendix will present the storage format and contents of the vector database used in our work. The database is divided into two sections: annotated input cards and function documentation.

An example of annotated input card, which were automatically generated using an LLM, is shown below (some of the content has been omitted due to space limitations):
\label{sec:data}
\begin{figure*}
\begin{lstlisting}[]
"input_card_name": "solid_mechanics_beam_static_torsion_1.i",
"overall_description": "This input card simulates the torsion of a beam fixed at one end with a moment applied at the other end. The simulation involves a 1D mesh of 10 elements representing the beam, with boundary conditions that fix all displacements and rotations at the left end. A constant moment of 5 Nm is applied at the right end, and the material properties are defined, including Young's modulus and Poisson's ratio. The simulation is transient, allowing for time-dependent analysis of the beam's response to the applied moment. The main applications used in this input card are the SolidMechanics for modeling the beam's mechanical behavior, the GeneratedMesh for creating the mesh, and the DirichletBC for applying boundary conditions.",
"annotated_input_card": "# Torsion test with automatically calculated Ix

# A beam of length 1 m is fixed at one end and a moment  of 5 Nm
# is applied along the axis of the beam.
# G = 7.69e9  # Shear modulus of the material in Pascals
# Ix = Iy + Iz = 2e-5  # Moment of inertia about the x-axis, calculated as the sum of the moments of inertia about the y and z axes
# The axial twist at the free end of the beam is:
# phi = TL/GIx = 3.25e-4  # Angle of twist at the free end of the beam in radians

[Mesh]  # Mesh block defining the geometry of the simulation
  type = GeneratedMesh  # Type of mesh to be generated
  dim = 1  # Dimension of the mesh (1D for beam)
  nx = 10  # Number of elements in the x direction
  xmin = 0.0  # Minimum x-coordinate of the mesh
  xmax = 1.0  # Maximum x-coordinate of the mesh
  displacements = 'disp_x disp_y disp_z'  # Displacement variables to be solved
[]

[Physics/SolidMechanics/LineElement/QuasiStatic]  # Physics block for solid mechanics in a quasi-static context
  [./block_all]  # Block for all elements in the mesh
    add_variables = true  # Indicates that variables should be added to the simulation
    displacements = 'disp_x disp_y disp_z'  # Displacement variables
    rotations = 'rot_x rot_y rot_z'  # Rotation variables for the beam

    # Geometry parameters
    area = 0.5  # Cross-sectional area of the beam
    Iy = 1e-5  # Moment of inertia about the y-axis
    Iz = 1e-5  # Moment of inertia about the z-axis
    y_orientation = '0.0 1.0 0.0'  # Orientation of the beam in the y-direction
    block = 0  # Block ID for the geometry
  [../]  # End of block_all
[]
...
</Input Card>"
\end{lstlisting}
% \caption{Annotated Moose input file.}
% \label{FIG:Annotated Moose input file}
\end{figure*}

An example of the Moose function is shown below (some of the content has been omitted due to space limitations):

\begin{figure*}
\begin{lstlisting}[]
"App_name": "ConvectiveFluxBC",
"description": "The `ConvectiveFluxBC` boundary condition determines the value on a boundary based upon
the initial and final values, the flux through the boundary and the duration of exposure.",
"document": "# "Determines boundary values via the initial and final values, flux, and exposure duration"
absolute_value_vector_tags = (no_default)     # "The tags for the vectors this residual object should
# fill with the absolute value of the residual contribution"
# Group: "Tagging"
boundary                   = (required)       # "The list of boundary IDs from the mesh where this
# object applies"
# Group: ""
control_tags               = (no_default)     # "Adds user-defined labels for accessing object parameters
# via control logic."
# Group: "Advanced"
diag_save_in               = (no_default)     # "The name of auxiliary variables to save this BC's
# diagonal jacobian contributions to.  Everything
# about that variable must match everything about
# this variable (the type, what blocks it's on, etc.)"
# Group: "Advanced"
displacements              = (no_default)     # "The displacements"
...
use_displaced_mesh         = 0                # "Whether or not this object should use the displaced
# mesh for computation.  Note that in the case this
# is true but no displacements are provided in the
# Mesh block the undisplaced mesh will still be used."
# Group: "Advanced"
use_interpolated_state     = 0                # "For the old and older state use projected material
# properties interpolated at the quadrature points.
# To set up projection use the ProjectedStatefulMaterialStorageAction."
# Group: ""
variable                   = (required)       # "The name of the variable that this residual object
# operates on"
# Group: ""
vector_tags                = nontime          # "The tag for the vectors this Kernel should fill"
# Group: "Tagging""
\end{lstlisting}
% \caption{Moose function.}
% \label{FIG:Moose function}
\end{figure*}

\section*{Declaration of competing interest}
The authors declare that they have no known competing financial interests or personal relationships that
could have appeared to influence 
the work reported in this paper.
\section*{Data availability}
The code for MooseAgent has been made open source and is available at
\href{https://github.com/taozhan18/MooseAgent}{https://github.com/taozhan18/MooseAgent}.
\bibliographystyle{plain}
\bibliography{references}

\end{document}